\documentclass[letterpaper, 10 pt, conference]{ieeeconf}  % Comment this line out if you need a4paper

\IEEEoverridecommandlockouts                              % This command is only needed if 
\usepackage{graphics} % for pdf, bitmapped graphics files
\usepackage{graphicx} 
\title{\LARGE \bf
 Challenges and Research Directions from the Operational Use of a Machine Learning Damage Assessment System via Small Uncrewed Aerial Systems at Hurricanes Debby and Helene}

\author{Thomas Manzini$^{*1}$, Priyankari Perali$^{*1}$, Dr. Robin R. Murphy$^{1}$, and David Merrick$^{2}$ 
\thanks{*Indicates equal contribution by authors}% <-this % stops a space
\thanks{$^{1}$Thomas Manzini, Priyankari Perali, and Dr. Robin R. Murphy are with the Department of Computer Science at Texas A\&M University. 
        435 Nagle St, College Station, TX, United States, 77843.
        {\tt\small tmazini@tamu.edu, perali@tamu.edu, robin.r.murphy@tamu.edu}}%
\thanks{$^{2}$David Merrick is with the Center for Disaster Risk Policy at Florida State University,
        113 Collegiate Loop, Tallahassee, FL, United States, 32306
        {\tt\small dmerrick@em.fsu.edu}}%
}

\begin{document}

\maketitle
\thispagestyle{empty}
\pagestyle{empty}

%%%%%%%%%%%%%%%%%%%%%%%%%%%%%%%%%%%%%%%%%%%%%%%%%%%%%%%%%%%%%%%%%%%%%%%%%%%%%%%%
\begin{abstract}

This paper details four principal challenges encountered with machine learning (ML) damage assessment using small uncrewed aerial systems (sUAS) at Hurricanes Debby and Helene that prevented, degraded, or delayed the delivery of data products during operations and suggests three research directions for future real-world deployments. 
%This paper outlines four principal challenges encountered during the deployment of a small uncrewed aerial systems (sUAS)--based machine learning (ML) damage assessment system at Hurricanes Debby and Helene that prevented, degraded, or delayed the delivery of ML data products during operations and suggests three research directions for future real-world deployments. 
The presence of these challenges is not surprising given that a review of the literature considering both datasets and proposed ML models suggests this is the first sUAS-based ML system for disaster damage assessment actually deployed as a part of real-world operations. 
The sUAS-based ML system was applied by the State of Florida to Hurricanes Helene (2 orthomosaics, 3.0 gigapixels collected over 2 sorties by a Wintra WingtraOne sUAS) and Debby (1 orthomosaic, 0.59 gigapixels collected via 1 sortie by a Wintra WingtraOne sUAS) in Florida.
The same model was applied to crewed aerial imagery of inland flood damage resulting from post-tropical remnants of Hurricane Debby in Pennsylvania (436 orthophotos, 136.5 gigapixels), providing further insights into the advantages and limitations of sUAS for disaster response.
%This work provides a case study of the deployment of an ML building damage assessment algorithm on sUAS imagery as part of state agencies’ response to Hurricanes Helene (2 orthomosaics, 3.0 gigapixels collected N flights by a Wintra WingtraOne sUAS) and Debby (1 orthomosaic, 0.59 gigapixels collected over N flights by an UNKNOWN sUAS) in Florida, and compares the application of that same model to crewed aerial imagery of inland flood damage resulting from post-tropical remnants of Hurricane Debby in Pennsylvania (3 orthomosaics and 436 orthophotos, 136.5 gigapixels). This work describes the imagery collected, the infrastructure used to deploy the ML model in practice, and identifies four challenges encountered and their impact YET MORE REDUNDANCY. 
The four challenges (variation in spatial resolution of input imagery, spatial misalignment between imagery and geospatial data, wireless connectivity, and data product format) 
%impacted potential operational value by either preventing, degrading, or delaying the delivery of data products during operations. NOTE: REDUNDANT
lead to three recommendations that specify research needed to improve ML model capabilities to accommodate the wide variation of potential spatial resolutions used in practice, handle spatial misalignment, and minimize the dependency on wireless connectivity.
%and data compute partitioning (\textbf{LIST HERE - ODD TO DO IT FOR CHALLENGES BUT NOT RECOMMENDATIONS}). - Added in recommendations
These recommendations are expected to improve the effective operational use of sUAS and sUAS-based ML damage assessment systems for disaster response.
\end{abstract}

%%%%%%%%%%%%%%%%%%%%%%%%%%%%%%%%%%%%%%%%%%%%%%%%%%%%%%%%%%%%%%%%%%%%%%%%%%%%%%%%

\section{INTRODUCTION}
%TOM: Valid yellow highlighter test
%%% FEEDBACK: NEED TO FOCUS ON ROBOTS AND THE POINT OF THE PAPER INSTEAD OF TWO PARAGRAPHS OF LESSeR RELEVANT INFO. 

This work reports on the first known deployment of a small uncrewed aerial system (sUAS) using machine learning (ML) for damage assessment after a disaster, where the sUAS and data products were operationally used by emergency managers during the response. It also compares the application of the same model to crewed aircraft imagery. The use of aerial imagery in disaster management has become widespread as aerial assets are capable of viewing areas that are inaccessible or hazardous to humans \cite{LozanoIJDRR:23}, however, they frequently produce unmanageable quantities of data. Unfortunately, this results in delayed or incomplete decisions as decision makers attempt to cope with the large volume of data, thus motivating machine learning (ML) approaches to rapidly analyze and label aerial imagery \cite{AnkerAOR:19,garcia2021computer}. 

\begin{figure}
  \centering
  \includegraphics[width=0.95\columnwidth]{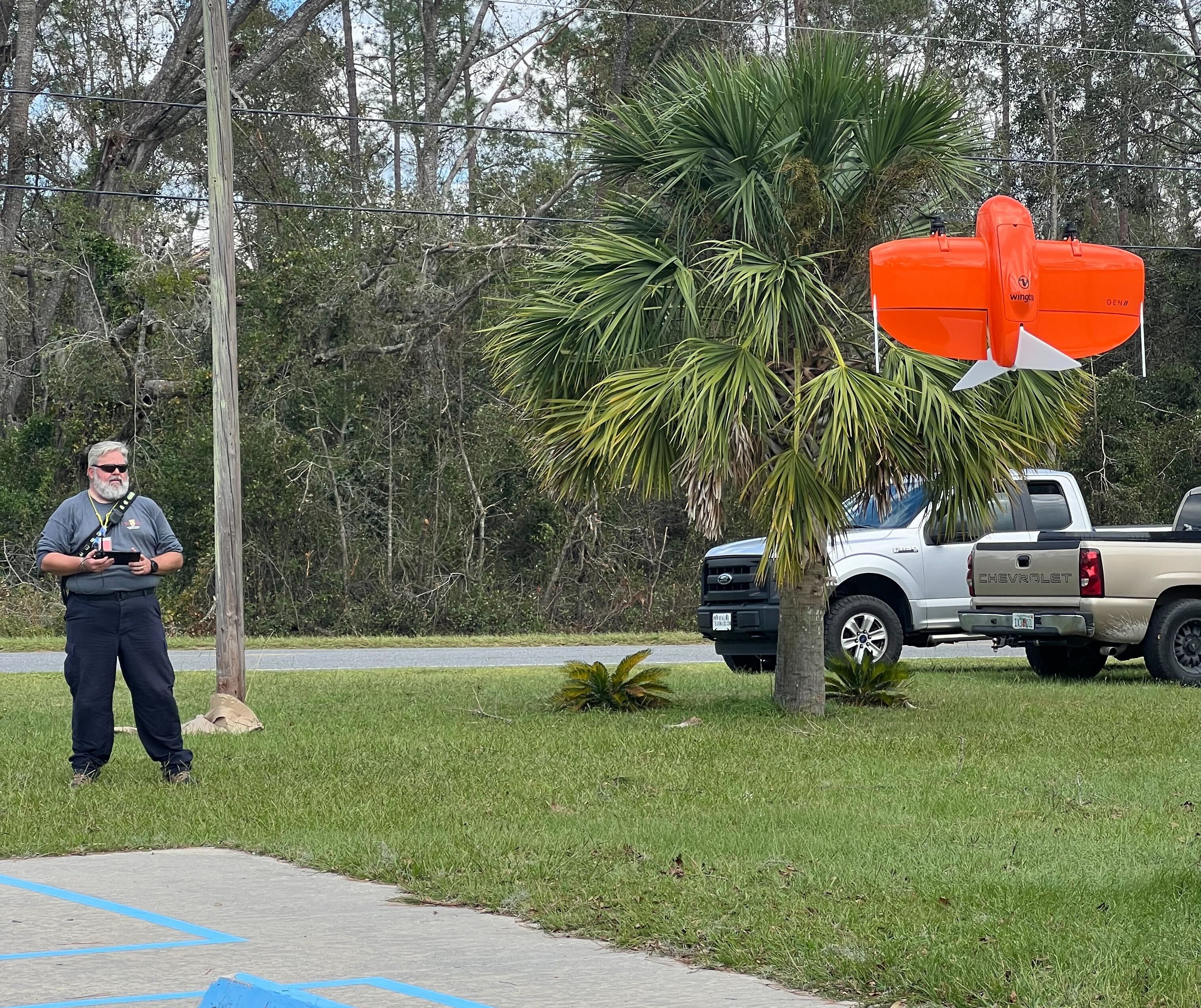}
  \caption{Example of a sUAS used in response to Hurricane Helene.}
  \label{fig:uas_in_flight}
\end{figure}

The application of the sUAS-based ML damage assessment system at Hurricanes Debby and Helene, Category 1 and 4 storms, respectively, that made landfall in the United States during the 2024 Hurricane Season\cite{noaa2024hurricanes} 
%appears to be the first known instance in which 
%ML-based  system data products from sUAS imagery were given to emergency managers during the response for operational use. 
was performed by two members of the NSF AI Institute for Societal Decision Making. 
Florida State University’s (FSU) Center for Disaster Risk Policy conducted sUAS missions on behalf of local and state agencies having jurisdiction as part of the Florida State Emergency Response Team FL-UAS1 task force, while Texas A\&M provided the sUAS-based ML damage assessment system. In addition, the Pennsylvania Emergency Management Agency (PEMA) requested Texas A\&M apply the same system to inland flooding caused by the post-tropical remnants of Hurricane Debby using imagery from crewed aircraft with similar resolution to sUAS. Fig.~\ref{fig:uas_in_flight} shows a sUAS used by FSU at Hurricane Helene.

%NOTE: TO GET THIS FIGURE ON PAGE 2 
%%% NOTE WHY ON PAGE 2 SINCE THE MATERIAL IT IS ABOUT IS NOT UNTIL PAGE 3?
\begin{figure*}[t]
  \centering
  \includegraphics[width=0.92\textwidth]{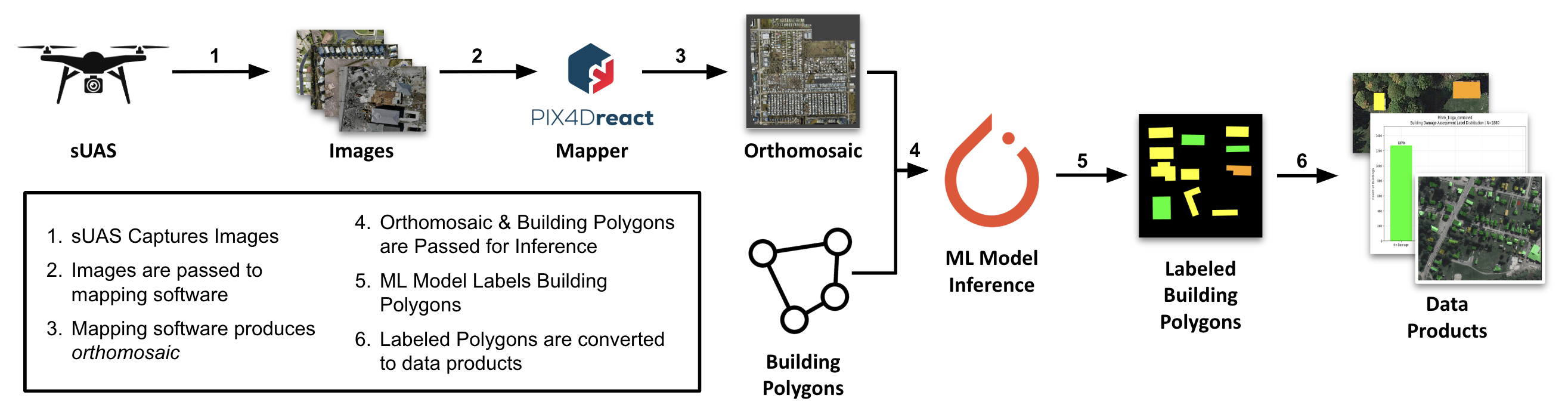}
  \caption{Visual of sUAS-based ML Damage Assessment System and inference pipeline deployed at Hurricanes Debby and Helene. }
  \label{fig:pipeline}
\end{figure*}

%TOM: Valid yellow highlighter test

%This work reports on what appears to be the first deployment of one such sUAS-based ML damage assessment system in a real-world setting with sUAS while also comparing the application of the same model to crewed aircraft imagery.
%In this case, imagery was collected by sUAS and crewed aircraft, and model outputs were delivered to emergency managers for review.
%Through the deployment of this system, the authors REALLY USINGS "AUTHORS" instead of just saying Foun unexpected challenges were encountered? encountered four unexpected challenges that either degraded, delayed, or prevented the delivery of the ML system data products to emergency managers. These challenges motivate this work to alert the robotics and human-robot interaction (HRI) community to these problems in hopes of facilitating the improved integration for real-world disaster response. 
% This work serves to alert the community to these problems in hopes of directing the communities efforts towards problems that will lead to improvements in the real-world use of such systems.

The core contribution of this work is an analysis of the deployment of such a system in practice; it is not a study of the accuracy or correctness of the sUAS-based ML damage assessment system. 
 Sec.~\ref{sec:related_work} discusses previous operational use of sUAS. 
 Sec.~\ref{sec:suas_ml_deployment_hurricane_helene_debby} details the sUAS and crewed aircraft operations during the response to Hurricanes Debby and Helene (Sec. \ref{sec:suas_crewed_aircraft_operations}), followed by a description of the sUAS-based ML damage assessment system (Sec. \ref{sec:ml_model_inference_pipeline}).
Sec.~\ref{sec:findings_from_hurricane_helene_debby} describes previously unreported challenges and considerations that limit the applicability and effectiveness of the sUAS-ML system: spatial resolution, spatial misalignment, wireless connectivity, and data product format.
In response to these challenges, this paper concludes in Sec.~\ref{sec:conclusion_recommendations} with three recommendations for the robotics  
%% WHERE THE HECK DID HRI COME FROM? THIS IS OUT OF NOWHERE FROM THE INTRO AND NOT MOTIVATED
%and HRI 
community to facilitate improved applicability and effectiveness of sUAS-based ML systems for real-world deployments. 
As such, the paper is expected to alert both the robotics and ML communities to current challenges requiring further research and suggest recommendations for the future.

%TOM: Valid yellow highlighter test

% THIS IS NOT INTRODUCTION MATERIAL. THE TWO SUBSECTIONS GO INTO SECTION III-HWERE THE APPROACH WOULD NORMALLY BE. 
% FINDINGS SHOULD BE SECTION IV, WHERE EXPERIMENTS AND FINDINGS WOULD NORMALLY BE. -- Moved to Section III: sUAS and ML model deployment at Hurricanes Debby and Helene

\begin{table*}[]
\caption{Spatial Resolution Statistics for the imagery collected during Hurricanes Debby and Helene ordered by capture date.}
\label{tab:ortho_details}
\begin{tabular}{|c|c|c|c|c|c|c|c|}
\hline
\textbf{Event}                                                     & \textbf{\begin{tabular}[c]{@{}c@{}}Capture\\ Date\end{tabular}} & \textbf{\begin{tabular}[c]{@{}c@{}}Mission\\ Location\end{tabular}} & \textbf{\begin{tabular}[c]{@{}c@{}}sUAS\\ Model\end{tabular}} & \textbf{Sorties} & \textbf{\begin{tabular}[c]{@{}c@{}}Ground Sample\\ Distance\end{tabular}} & \textbf{\begin{tabular}[c]{@{}c@{}}Transmission\\ Size (GB)\end{tabular}} & \textbf{\begin{tabular}[c]{@{}c@{}}Imagery Format Delivered \\ From The Field\end{tabular}} \\ \hline
\begin{tabular}[c]{@{}c@{}}Hurricane Debby\\ (sUAS)\end{tabular}   & 2024-08-06                                                      & \begin{tabular}[c]{@{}c@{}}Branford\\ FL\end{tabular}               & \begin{tabular}[c]{@{}c@{}}Wingtra\\ WingtraOne\end{tabular}  & 1                & 4.51cm/px                                                                 & 14.10                                                                     & Raw Images                                                                                  \\ \hline
\begin{tabular}[c]{@{}c@{}}Hurricane Debby\\ (Crewed)\end{tabular} & 2024-08-16                                                      & \begin{tabular}[c]{@{}c@{}}Tioga County\\ PA\end{tabular}           & N/A                                                           & 1                & 9-11cm/px                                                                 & 456.0                                                                     & Orthomosaic                                                                                 \\ \hline
\begin{tabular}[c]{@{}c@{}}Hurricane Helene\\ (sUAS)\end{tabular}  & 2024-09-27                                                      & \begin{tabular}[c]{@{}c@{}}Mayo\\ FL\end{tabular}                   & \begin{tabular}[c]{@{}c@{}}Wingtra\\ WingtraOne\end{tabular}  & 1                & 1.65cm/px                                                                 & 7.01                                                                      & Orthomosaic                                                                                 \\ \hline
\begin{tabular}[c]{@{}c@{}}Hurricane Helene\\ (sUAS)\end{tabular}  & 2024-09-28                                                      & \begin{tabular}[c]{@{}c@{}}Dekle Beach\\ FL\end{tabular}            & \begin{tabular}[c]{@{}c@{}}Wingtra\\ WingtraOne\end{tabular}  & 1                & 25.3cm/px                                                                 & 0.015                                                                     & Orthomosaic                                                                                 \\ \hline
\end{tabular}
\end{table*}

\section{RELATED WORK}
\label{sec:related_work}

%TOM: Valid yellow highlighter test
The operational use of sUAS at disasters has been documented for at least 16 disasters. However, none of these deployments appear to have used a sUAS-based ML damage assessment system despite notable theoretical efforts. Without sUAS-based ML systems being used in practice, there is a lack of understanding of the challenges such systems may face when deployed for real-world use. 
%Furthermore, it is unclear what considerations should be made when developing future systems to ensure that they improve the operational benefit for emergency managers and decision-makers.

sUAS were deployed by emergency management agencies at 
Hurricanes Katrina \cite{murphy2008crew}, Harvey \cite{fernandes2018quantitative}, Michael \cite{fernandes2019quantitative}, Ian \cite{manzini2023quantitative}, Ike \cite{murphy2009robot}, Wilma \cite{murphy2008cooperative}, the Kilauea Volcano Eruption \cite{adams2018use}, the Surfside Condo Collapse \cite{rao2022analysis}, Hurricanes Florence, Irma, and Sally \cite{manzini2023quantitative}; Hurricanes Idalia, Ida, Laura, the Mayfield Tornado Outbreak, and the Musset Bayou Fire in \cite{manzini2024crasar}.
%%% WHAT THE HECK DOES THIS MEAN? THEY ARE CITED AND WE KNOW THEY WERE USED..
%"there is evidence that" 
sUAS were involved in Hurricane Dorian \cite{cheng2021dorianet} as well as Cyclone Idai and an unspecified earthquake in Nepal \cite{merkle2023drones4good}, but it is unclear as to whether emergency managers at those events had tasked sUAS and used the imagery or if the sUAS were used in an ad hoc role. 
Eight surveys of the ML and robotics literature discuss sUAS-based ML systems for disaster damage assessment \cite{al2024integrating, kerle2019uav, daud2022applications, linardos2022machine, kyrkou2022machine, krichen2023advances, chamola2020disaster, iqbal2023drones}, but none appear to have been used operationally or delivered model outputs to decision makers.

%TOM: Valid yellow highlighter test
% YOU CAN"T SAY THIS WITHOUT EXPLAINING WHAT LITERATURE.
%After a review of the literature, including eight surveys of the literature by the community that focus at least partially on sUAS-based ML systems for disaster damage assessment \cite{al2024integrating, kerle2019uav, daud2022applications, linardos2022machine, kyrkou2022machine, krichen2023advances, chamola2020disaster, iqbal2023drones}, there are appear to be no instances of such damage assessment systems being used in practice with sUAS.

%TOM: Valid yellow highlighter test
%% REPETITIVE AND OUT OF PLACE
%This work closes the gap within the literature by documenting the challenges and considerations for sUAS-based ML systems observed with the deployment of such systems at Hurricanes Debby and Helene.
%In response to the challenges, this work provides recommendations to improve the operational benefit of sUAS and sUAS-based ML systems used within disaster response for emergency managers and decision-makers.

\section{SUAS AND ML MODEL DEPLOYMENT AT HURRICANES DEBBY \& HELENE}
\label{sec:suas_ml_deployment_hurricane_helene_debby}

The approach taken in this paper relies on analyzing the data and direct observations captured during the response to Hurricanes Debby and Helene. In those two events, an sUAS-based ML model developed by Texas A\&M was deployed to accelerate the assessment of building damage from i) sUAS missions conducted in Florida by Florida State University and ii) crewed missions in Pennsylvania by PEMA that produced similar image resolution. %DO WE HAVE A CITATION FOR THIS - TOM: Unfortunately, not yet
The imagery captured by the deployments is detailed in Table~\ref{tab:ortho_details}, and Fig.~\ref{fig:pipeline} summarizes the sUAS-based ML damage assessment system and inference pipeline, both detailed below. 

%The remainder of this section with first detail the sUAS and crewed aircraft operation, followed by a description of the ML model.

\subsection{sUAS \& Crewed Aircraft Operations}
\label{sec:suas_crewed_aircraft_operations}
%TOM: Valid yellow highlighter test

In Florida, during the response to Hurricanes Debby and Helene, sUAS missions were not directed with the intention of using the sUAS-based ML system but instead were directed by agencies having jurisdiction and their operational objectives. The use of the ML system was secondary to those objectives, and its application was opportunistic and based on the availability of imagery that could be used with the inference pipeline described in Sec. \ref{sec:ml_model_inference_pipeline}. Of the sUAS missions conducted by FL-UAS1 during Hurricane Helene, two missions were conducted that collected data that could be used as input to the inference pipeline. During the response to Hurricane Debby, one mission was conducted, which resulted in imagery that was passed to the authors. All missions were flown by a Wintra WingtraOne sUAS. The complete list of the sUAS missions that resulted in imagery that was passed to the inference pipeline is shown in Table \ref{tab:ortho_details}.

In Pennsylvania, PEMA directed a local aerial imagery provider to collect crewed aerial imagery of Tioga County. All imagery was collected in a single sortie, and the primary objective of this sortie was to capture imagery of the Tioga River following the post-tropical remnants of Hurricane Debby, which had been the source of the flooding. This sortie resulted in the capture of 456GB of aerial imagery representing 436 orthophotos and 136.5 gigapixels of imagery. Upon termination of the sortie, the imagery was transferred to PEMA and then to the authors for analysis.

\subsection{sUAS-Based ML Damage Assessment System}
\label{sec:ml_model_inference_pipeline}
%TOM: Valid yellow highlighter test
The sUAS-based ML building damage assessment system takes raw imagery collected from sUAS and provides labels for buildings in the imagery according to the Joint Damage Scale \cite{gupta2019xbd}, as shown in Fig. \ref{fig:pipeline}. Functionally, this system is an inference pipeline which consists of six steps. 
% sUAS captures imagery, raw images are passed to mapping software, orthomosaic generation, orthomosaic and building polygons are passed to the model, model infers damage labels, and damage labels conversion to data products. 
First, during the response, the sUAS are flown to capture aerial imagery of an impacted area. The raw imagery collected is passed to mapping software (e.g., Pix4D and AgiSoft). The mapping software proceeds to ``stitch" the raw imagery together into a \textit{georectified orthomosaic}. A georectified orthomosaic is a collection of individual images combined to form a map where each pixel has a latitude and longitude. At this point, the georectified orthomosaic is a standalone data product that is typically transmitted to emergency managers and decision-makers for inspection.
Next, the resulting georectified orthomosaic and \textit{a priori} building polygons are passed to the ML model. A priori building polygons are existing spatial data provided by datasets like Microsoft Building Footprints \cite{MicrosoftBuildingFootprints}. The sUAS-based ML model infers damage labels for each of the building polygons provided based on the orthomosaic imagery. 
The model that was deployed was an Attention UNet \cite{oktay2018attention} that had been trained on the CRASAR-U-DROIDs dataset \cite{manzini2024crasar}.
Lastly, labels were converted to data products (e.g., GEOJSON/KML files, and counts of damaged buildings) for emergency managers and decision-makers to view.

\section{FINDINGS}
\label{sec:findings_from_hurricane_helene_debby}
%TOM: Valid yellow highlighter test

Four operational challenges were identified, which either led to the failure, delay, or degradation of ML model outputs.
These were: the wide variation in spatial resolution, 
% \textbf{IS THIS TRUE FOR JUST SUAS? THE PEMA STUFF IS FOR COMPARISON NOT MAINSTREAM,} -- This holds for both. but the failure within spatial resolution was during the response for helene in Florida
non-uniform spatial misalignment, limited wireless connectivity, and insufficient data product formats. 
% \textbf{ARE THESE IN THE ORDER OF IMPORTANCE?} -- These are ordered based on their observed impact (failure, degradation, delay)
Each challenge is discussed in detail below and ordered by observed impact.

%PROVIDE MORE USEFUL SUBSECTION HEADINGS - Changed subsection heading to be more useful

\subsection{Wide Variation in Spatial Resolution}
\label{sec:spatial_resolution}

\begin{figure}[]
  \centering
  \includegraphics[width=0.9\columnwidth]{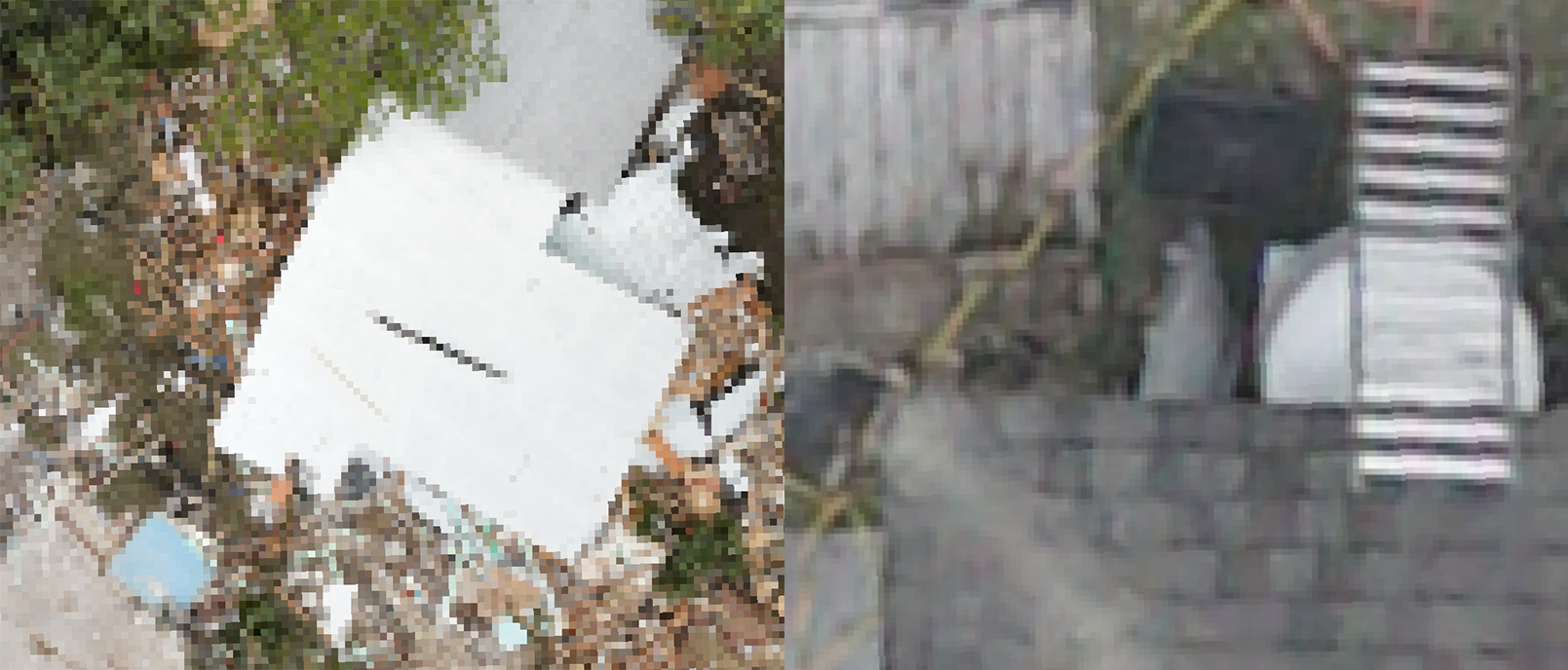}
  \caption{Comparison of spatial resolution of imagery captured during Hurricane Helene. Both images are presented at the same pixel scale. [Left] Imagery from Dekle Beach, FL, captured at 23.5 cm/px showing a partially collapsed building. [Right] Imagery from Mayo, FL, captured at 1.65cm/px showing the corner of a roof.}
  \label{fig:spatial_resolution}
\end{figure}

%TOM: Valid yellow highlighter test
Imagery ground sample distances (GSDs) were found to vary between 1.65cm/px and 25.3cm/px, representing more than one order of magnitude difference in imagery scales. A visual depiction of this difference in GSDs is shown in Fig. \ref{fig:spatial_resolution}. Notice that in Fig. \ref{fig:spatial_resolution}, the imagery on the right has a high enough resolution to discern individual roof tiles. However, the imagery on the left is 15x lower spatial resolution than the left, representing a far lower resolution than the data the sUAS-based ML model was trained on. This wide variety of spatial resolutions suggests that future ML models need to be capable of consuming a wider variety of spatial imagery than previously thought.
While performance degradations as a result of scale variation has been reported in the literature \cite{manzini2025now, maiti2022effect, vargas2019correcting}, and techniques have been proposed to manage such issues \cite{reed2023scale}, the specific challenge found in this work is the large degree to which imagery scales vary and the motivation for the variation. 

%TOM: Valid yellow highlighter test
Interestingly, in these operations, the variation in spatial resolution was not principally the result of variations in the sUAS model, or sensor, but instead was the result of the limited wireless connectivity that was observed in the field (see Sec. \ref{sec:wireless_connectivity}). In response to Hurricane Helene, the Dekle Beach, FL, sUAS imagery was downsampled and transmitted at a lower resolution to minimize the delay of imagery to emergency managers and decision-makers. The complete, full-resolution images associated with this orthomosaic would not be delivered to all authors until after the sUAS field teams had demobilized. This finding suggests that ML models must accommodate a wider variety of spatial resolutions than those based on known flight altitudes and sensors, as imagery resolution cannot be guaranteed.

%TOM: Valid yellow highlighter test
This variation in GSDs accounted for substantial differences between model outputs and, in the case of the downsampled Dekle Beach, FL orthomosaic, resulted in model outputs that were deemed invalid by the authors and not transmitted for distribution. 
After the demobilization of sUAS field teams, the authors were permitted access to the full-resolution imagery from the Dekle Beach, FL mission. When comparing the labels generated by the model, a label disagreement rate of 82.6\% was observed between the buildings in the full-resolution orthomosaic and the transmitted orthomosaic.
This label disagreement indicates that spatial resolution needs to be explicitly managed.

\subsection{Non-Uniform Spatial Misalignment}
\label{sec:spatial_misalignment}
\begin{figure}[t]
  \centering
  \includegraphics[width=0.9\columnwidth]{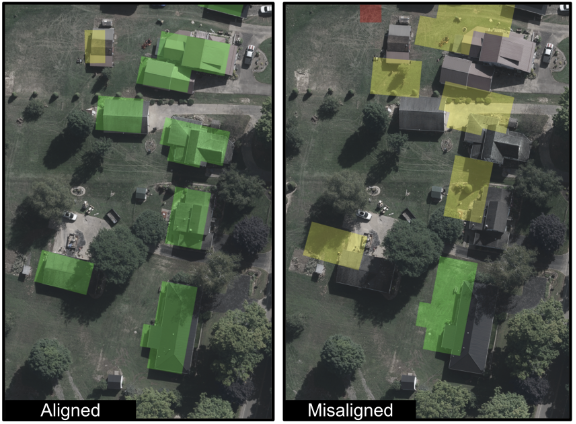}
  \caption{sUAS-based ML model outputs with aligned and misaligned a priori building polygons from deployment at Tropical Storm Debby in Tioga County, Pennsylvania. [Left] Aligned a priori building polygons with damage labels passed to the sUAS-based ML model. [Right] Misaligned a priori building polygons with damage labels passed to the sUAS-based ML model. Damage labels are colored as follows: no damage (green), minor damage (yellow), and destroyed (red). Note the label disagreement for six out of the seven damage labels between examples.}
  \label{fig:misalignment}
\end{figure}

%TOM: Valid yellow highlighter test
Spatial misalignment contributed to model degradations and delayed data-to-decision times in practice. 
Spatial misalignment refers to the misalignment between a priori spatial data and imagery \cite{manzini2024non}. A visual example of the spatial misalignment and model degradations is shown in Fig. \ref{fig:misalignment}, where misaligned a priori building polygons provided to the sUAS-based ML model resulted in 6 out of 7 labels disagreeing when compared to when aligned a priori building polygons were provided. 
% \textbf{EXPLAIN MORE THIS READS AS A "GO FIGURE IT OUT YOURSELF" } 
Misalignment can be manually corrected, though doing so delays the delivery of data products.% On the contrary, misalignment left uncorrected poses a potential for variations in model predictions. 

%TOM: Valid yellow highlighter test
Correcting the misalignment present in sUAS orthomosaics required human intervention, thus increasing data-to-decision times. 
Manual correction of misalignment was done for the data collected at Dekle Beach, FL, for Hurricane Helene. Following the process within \cite{manzini2024non} for manual adjustments, 13 adjustments were made by the authors to correct 46 building polygons' misalignment. This corrected an average of 30.56px and 141.72 degrees of misalignment. 
%While adjustments were made to correct the misalignment, this increased data to decision times because the misalignment had to be corrected before running the model. 
%Due to this, manual correction was not possible for any of the other deployments.

%TOM: Valid yellow highlighter test
Misalignment contributes to variations in model predictions and likely degrades model performance if not corrected. 
Consider, as an example, the Dekle Beach orthomosaic. Misalignment was observed in this orthomosaic in addition to lower than expected resolution (see Sec. \ref{sec:spatial_resolution}). Once teams were demobilized and the full resolution was available, the misalignment was manually corrected. After aligned building polygons were available, the inference pipeline was run on the aligned and misaligned building polygons, and a 39.1\% rate of label disagreement was observed.
This label disagreement suggests that spatial misalignment has the potential to substantially degrade model performance.

% \subsection{Geographic Variability}
% \label{sec:geographic_variability}
% Geographic shifts in imagery from training data contributed to misclassifications of region-specific building structures and terrain. At Hurricane Debby, imagery collected from Tioga County, PA contained buildings with black roofs, varying structures of buildings, and varying terrain (e.g., valleys, mountainous areas). The sUAS-based ML model was trained with the CRASAR-U-DRIODs \cite{manzini2024crasar} dataset. While this dataset is the largest sUAS building damage assessment dataset, the building structures and terrain from Tioga County were not present in the dataset, presenting a potential distribution shift. When such building structures were presented to the model, the model provided spurious damage labels that were obviously incorrect. During the deployments at Hurricane Debby and Helene, if such incorrect damage labels were provided by the model, they were not given to decision-makers to avoid negative downstream effects. 

\subsection{Limited Wireless Connectivity}
\label{sec:wireless_connectivity}
%TOM: Valid yellow highlighter test
Limited wireless connectivity delayed when imagery could be transmitted for processing, thereby delaying the execution of the inference pipeline and, thus, the delivery of model outputs to emergency managers and decision-makers.
This was the case with all sUAS aerial imagery missions flown in response to Hurricanes Debby and Helene in Florida and for the crewed aerial imagery missions flown in response to Hurricane Debby in Pennsylvania. In all instances, the largest delays in the delivery of data products to emergency managers and decision-makers were related to the transmission of the data as opposed to delays related to the ML model or the generation of data products. In all instances, the time between when the authors were informed that data had been collected and when the data would be made available for inference was approximately one day. The apparent limited capability to transfer gigabyte-scale data from the field to inference compute was the single most impactful factor delaying the delivery of ML data products to emergency managers and decision-makers.

%TOM: Valid yellow highlighter test
In the case of the deployment of the ML model on imagery collected as a part of the sUAS response to Hurricanes Debby and Helene in Florida, imagery was delayed as a result of limited wireless connectivity in the field. This delay was two-fold, first consistent with previous deployments \cite{manzini2023wireless} sUAS teams had to rely on limited wireless connectivity through SpaceX StarLink and AT\&T FirstNet, and second because operational objectives took priority over the transmission of imagery via the available wireless connectivity. 
In one instance, during Hurricane Helene, an orthomosaic was physically transported to emergency managers, meaning there was no opportunity to run the damage assessment system until after sUAS field teams had demobilized.

%TOM: Valid yellow highlighter test
In the case of the deployment of the ML model on imagery collected as a part of the crewed aerial imagery missions flown in response to Hurricane Debby in Pennsylvania, the execution of the inference pipeline was delayed as a result of insufficient download speeds after delivery of the data by PEMA. In this instance, imagery was shared with the authors via an internet-connected cloud service that required all imagery to be downloaded via a web browser and transferred to the compute where the inference pipeline could be run. %TODO CONSIDER: Could this be read as a negative for PEMA? perhaps rephrase

\subsection{Insufficient Data Product Formats}
\label{sec:data_products_format}
%TOM: Valid yellow highlighter test
Spatial data products (GEOJSON and KML files) were not always sufficient to provide the necessary information to emergency managers and decision-makers. Initially, at Hurricane Debby, it was assumed that spatial data products would be easily comprehensible and preferred by emergency managers and decision-makers\cite{national2007successful}. However, operations in Florida during both Hurricanes Debby and Helene saw limited resources to visualize and comprehend such formats. 

%TOM: Valid yellow highlighter test
Expertise in interpreting Geographic Information System (GIS) data products may not always be available, especially if teams are working in wireless-denied environments within the disaster site. 
For instance, the sUAS operations during Hurricane Debby had access to one ``GIS specialist" at an Emergency Operations Center who quickly became task-saturated as disaster operations progressed.
With such specialists having limited bandwidth, spatial data products like KML and GEOJSONs were not easily comprehensible to individuals who were not familiar with them. As a result, at the guidance of emergency managers and decision-makers following Hurricane Debby, the inference pipeline was updated to return model outputs as CSV and PDF files, including street addresses, in addition to the aforementioned spatial data products. This was done so that the building-level outputs from the inference pipeline could be transferred to and interpreted on computers without GIS software.

\begin{figure}[t]
  \centering
  \includegraphics[width=0.9\columnwidth]{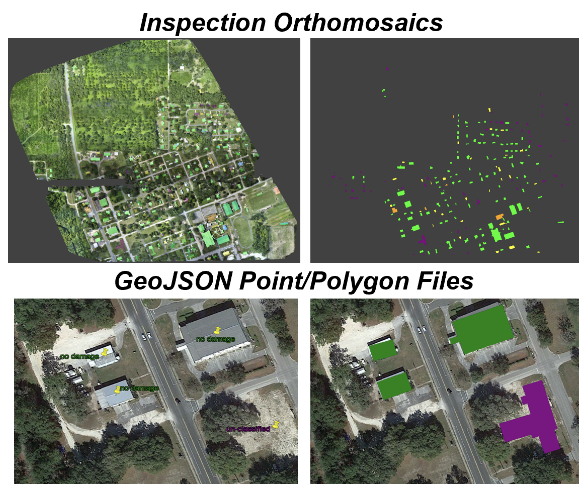}
  \includegraphics[width=0.9\columnwidth]{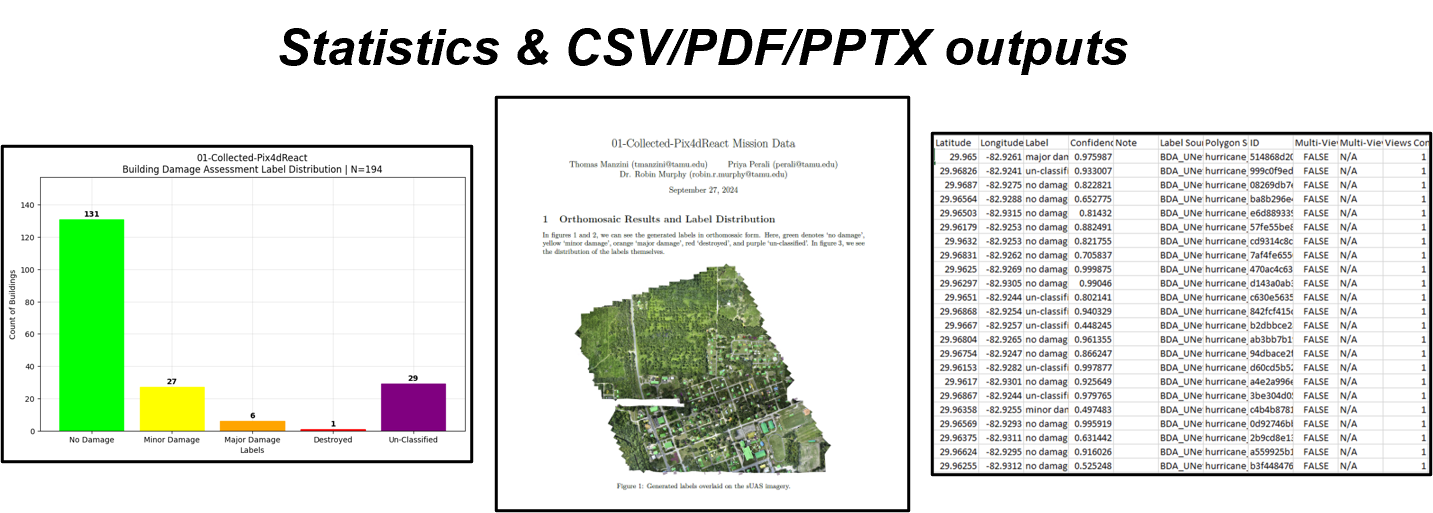}
    \caption{Examples of data products provided during Hurricane Debby and Helene. Inspection Orthomosaics, GeoJSON Point/Polygon Files, and Statistics (bottom-left bar graph) data products were provided during the response to Hurricane Debby. CSV/PDF/PPTX outputs were added to the data products during Hurricane Helene.}
  \label{fig:data_products}
\end{figure}

%TOM: Valid yellow highlighter test
At Hurricane Debby, the data product formats provided consisted of inspection orthomosaics, GEOJSON point and polygon files, and overall statistics. 
Visual examples of these data products are shown in Fig. \ref{fig:data_products}. The inspection orthomosaics consisted of two orthomosaics (top row of Fig. \ref{fig:data_products}).
% \textbf{REFER BACK TO THE FIGURE TO HELP THE READER. THIS IS DISJOINT.} 
First, the original imagery was overlaid with colored building polygons, with the colors representing the damage label output provided by the model (top left orthomosaic in Fig. \ref{fig:data_products}). The second orthomosaic contained only the colored building polygons and provided no imagery (top right orthomosaic in Fig. \ref{fig:data_products}). The GEOJSON point and polygon files contained the latitude and longitude points for each of the buildings and their respective damage labels provided by the model. These points were provided in KML and GEOJSON files, which could be viewed through GIS mapping tools (e.g., ArcGIS) (middle row of Fig. \ref{fig:data_products}). Lastly, the overall statistics provided consisted of bar charts for damage counts and confidence scores for each damage label, and JSON and CSV files for damage label counts (bottom row of Fig. \ref{fig:data_products}).

\section{CONCLUSION \& RECOMMENDATIONS}
\label{sec:conclusion_recommendations}
%TOM: Valid yellow highlighter test

The application of the sUAS-based ML building damage assessment system at Hurricanes Debby and Helene shows that such systems are valuable to emergency management and are on the verge of being ready for adoption. 
Four challenges were encountered in the deployment of the sUAS-based ML damage assessment system: variation in spatial resolution of input imagery, spatial misalignment between imagery and geospatial data, wireless connectivity, and data product format.  Based on these challenges,  three directions for future robotics and ML research into accommodating varying spatial resolutions, automating spatial misalignment, and minimizing dependency on wireless connectivity is expected to both add to the fundamental understanding of sUAS and to benefit the operational use of sUAS and sUAS-based ML damage assessment systems for decision-makers in disaster response. Each recommendation is detailed below.  

%NUMBER THE RECOMMENDATIONS. - Recommendation put into numbered list
\begin{enumerate}
    \item \textit{ML models for automated damage assessment must be built to accommodate the variety of spatial resolutions that could be reasonably produced from sUAS flights over disaster scenes}. As discussed in Sec. \ref{sec:spatial_resolution}, the wide variety of spatial resolutions that were produced during the response cannot be determined a priori. This is because imagery may be transmitted at varying resolutions in response to operational objectives or limitations stemming from wireless connectivity. Instead, ML systems must be built to handle the wide variety of spatial resolutions that could reasonably appear in practice, as opposed to a selection of spatial resolutions that sUAS typically produce.
    \item \textit{ML models for automated damage assessment need to gracefully handle the spatial misalignments present in sUAS imagery to provide accurate assessments of damage}. As discussed in Sec. \ref{sec:spatial_misalignment}, spatial misalignment either had to be manually corrected or tolerated due to strained resources, which either delayed data-to-decision times or posed a risk for degraded model performance. Without an automated approach to correcting the misalignment present within sUAS imagery, it is impractical to expect that misalignment will be manually corrected in the field, nor that decision-makers will be aware of how to interpret the potential for label disagreement within data products due to misalignment. Therefore, future deployments of sUAS-based ML models should develop and utilize an approach to correct the spatial misalignment within the sUAS imagery before providing damage assessments. 
    \item \textit{Wireless and robotics experts need to work together to create more suitable Cloud/Edge partitioning for machine learning in disasters to minimize dependence on wireless connectivity}. As discussed in Sec. \ref{sec:wireless_connectivity}, the lack of stable connection between sUAS teams in the field and the compute that was used to run the sUAS-based ML damage assessment inference pipeline was the single largest source of delays. This suggests that the sUAS-based ML damage assessment inference pipeline must be co-located with the sUAS team in the field to minimize the distance (both physical and virtual) that is required to transmit imagery for inference. For future deployments, teams anticipating the use of sUAS-based ML damage assessment systems should expect to deploy GPU-based compute into the field alongside sUAS field teams to remove delays associated with the wireless transmission of imagery.
\end{enumerate}

% \textit{sUAS disaster imagery datasets should be expanded to more locales to capture the variety of environments in which disasters may occur}. The performance degradations observed in Section \ref{sec:geographic_variability} suggest that as ML models are utilized in other locales they are hindered by geographical distribution shifts. As these systems become more integrated within disaster response operations, they must provide operational benefits in a variety of geographical and building structures. Without current datasets that capture a large variety of environments, the operational benefit for these systems will be limited. Therefore, future research should invest in developing benchmark datasets that are expanded to a larger variety of environments to reduce model performance degradation due to geographical distribution shifts. 

\section*{ACKNOWLEDGMENTS}
This work is supported by the AI Research Institutes Program funded by the National Science Foundation under the AI Institute for Societal Decision Making (NSF AI-SDM), Award No. 2229881, and under ``Datasets for Uncrewed Aerial System (UAS) and Remote Responder Performance from Hurricane Ian" Award No. 2306453. 
%GEEZ, this is repetitive Acknowledgments are given to
The authors thank the Florida State Emergency Response Team, FL-UAS1 task force, and the Pennsylvania Emergency Management Agency for their engagement and feedback.
%support THIS SOUNDS LIKE FINANCIAL of this work and willingness to involve us in their operations.

%%%%%%%%%%%%%%%%%%%%%%%%%%%%%%%%%%%%%%%%%%%%%%%%%%%%%%%%%%%%%%%%%%%%%%%%%%%%%%%%

%%%%%%%%%%%%%%%%%%%%%%%%%%%%%%%%%%%%%%%%%%%%%%%%%%%%%%%%%%%%%%%%%%%%%%%%%%%%%%%%

\bibliographystyle{IEEEtranS}
\bibliography{references}

\end{document}